\documentclass[letterpaper, 10 pt, conference]{ieeeconf}
\IEEEoverridecommandlockouts

\usepackage{xspace}
\usepackage{xcolor}
\usepackage{amsfonts}
\usepackage{graphics} %
\graphicspath{{figures/}}  %
\usepackage{epsfig} %
\usepackage{mathptmx} %
\usepackage{times} %
\usepackage{amsmath} %
\usepackage{amssymb}  %
\usepackage{ulem}
\usepackage{siunitx}
\usepackage{verbatim}
\usepackage{booktabs}
\usepackage{multirow}
\usepackage[space]{cite}
\usepackage[table]{xcolor} %
\definecolor{lightgreen}{RGB}{210,255,210}
\usepackage[colorlinks=true, linkcolor=blue, urlcolor=blue, citecolor=blue]{hyperref}

\usepackage[font=footnotesize]{caption}

\makeatletter
\let\origthebibliography\thebibliography
\def\thebibliography#1{%
  \origthebibliography{#1}%
  \scriptsize %
}
\makeatother

\newcommand{\method}{ROPA\xspace}
\newcommand{\ba}{\mathbf{a}}

\definecolor{figgray}{rgb}{0.2, 0.2, 0.2}
\definecolor{figgreen}{rgb}{0.180, 0.494, 0.196}
\definecolor{figblue}{rgb}{0.082, 0.396, 0.753}
\definecolor{figlightred}{rgb}{1.0, 0.149, 0.0}
\definecolor{figdarkblue}{rgb}{0.188, 0.392, 0.729}

\title{\LARGE \bf
\method: Synthetic Robot Pose Generation \\ for RGB-D Bimanual Data Augmentation
}

\author{
  Jason Chen, I-Chun Arthur Liu, 
  Gaurav S. Sukhatme, Daniel Seita \\
  University of Southern California
}

\begin{document}
\twocolumn[{%
\renewcommand\twocolumn[1][]{#1}%
\maketitle
\vspace{-2em}
\begin{center}
    \centering
    \captionsetup{type=figure}
    \includegraphics[width=1.0\textwidth]{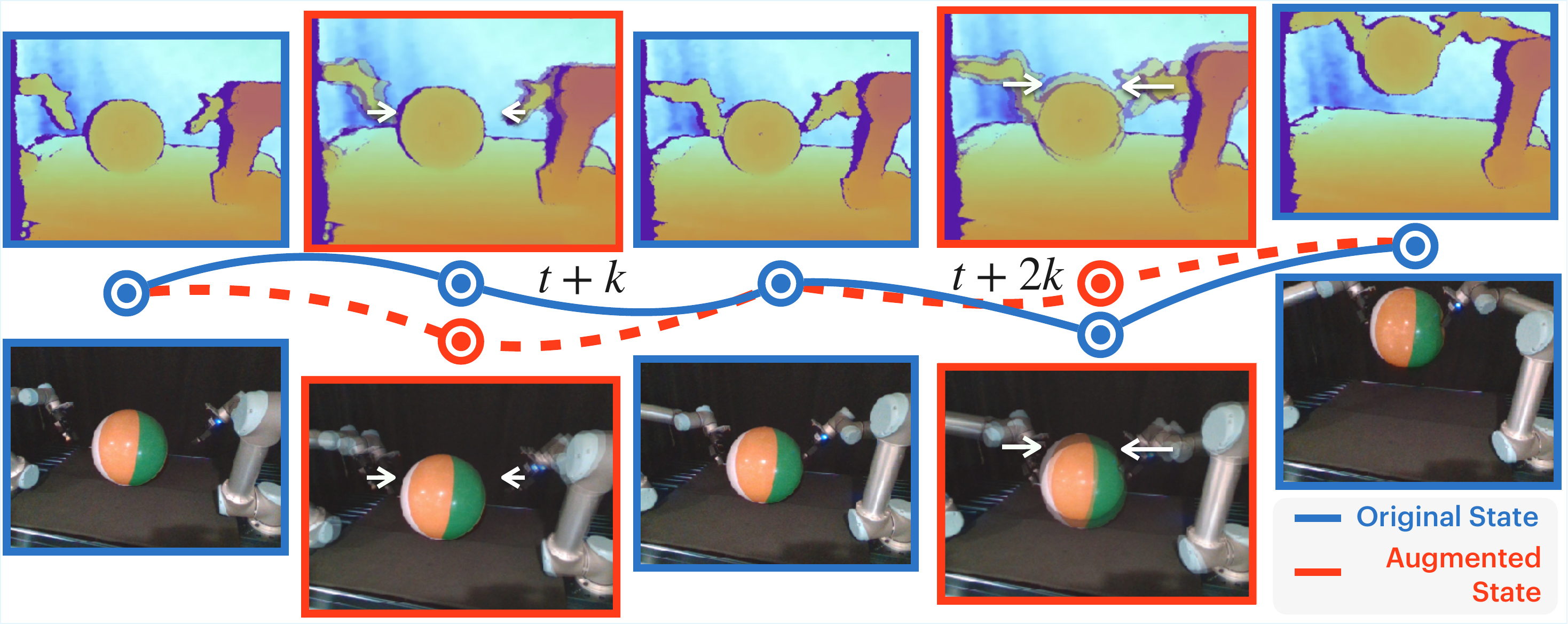}
    \vspace{-10pt}
    \captionof{figure}{
        \textbf{\method} performs offline data augmentation for bimanual imitation learning. White arrows indicate pose differences between the original and augmented images. \textcolor{figlightred}{Red regions} represent \method generated images and states every $k$ timesteps at $t+k$ and $t+2k$, while \textcolor{figdarkblue}{blue regions} show the original dataset. RGB and depth image pairs are captured at the same timesteps, with the top row displaying depth colormap and the bottom row showing standard RGB images.}

    \label{fig:teaser}
\end{center}%
}]
\thispagestyle{empty}
\pagestyle{empty}

\begin{abstract}
Training robust bimanual manipulation policies via imitation learning requires demonstration data with broad coverage over robot poses, contacts, and scene contexts. 
However, collecting diverse and precise real-world demonstrations is costly and time-consuming, which hinders scalability. Prior works have addressed this with data augmentation, typically for either eye-in-hand (wrist camera) setups with RGB inputs or for generating novel images without paired actions, leaving augmentation for eye-to-hand (third-person) RGB-D training with new action labels less explored. In this paper, we propose Synthetic \underline{Ro}bot \underline{P}ose Generation for RGB-D Bimanual Data \underline{A}ugmentation (\method), an offline imitation learning data augmentation method that fine-tunes Stable Diffusion to synthesize third-person RGB and RGB-D observations of novel robot poses. Our approach simultaneously generates corresponding joint-space action labels while employing constrained optimization to enforce physical consistency through appropriate gripper-to-object contact constraints in bimanual scenarios. We evaluate our method on 5 simulated and 3 real-world tasks. Our results across 2625 simulation trials and 300 real-world trials demonstrate that \method outperforms baselines and ablations, showing its potential for scalable RGB and RGB-D data augmentation in eye-to-hand bimanual manipulation.
Our project website is available at: \href{https://ropaaug.github.io/}{https://ropaaug.github.io/}.
\end{abstract}

\section{Introduction}
\begingroup
  \setlength{\skip\footins}{1pt}    %
  \setlength{\footnotesep}{2pt}     %
  \let\thefootnote\relax
    \footnotetext{%
    All authors are with the Thomas Lord Department of Computer Science at the University of Southern California, USA. Correspondence: \texttt{jchen567@usc.edu}.%
  }
\endgroup
Bimanual manipulation is critical for a wide range of daily tasks~\cite{Bimanual_Taxonomy_2022} such as lifting large objects~\cite{stepputtis2022bimanual,xie2020}, handling deformable objects~\cite{avigal2022speedfolding,grannen2023stabilize,maitin2010cloth,weng2021fabricflownet}, and opening containers~\cite{liu2024voxactb,twist_lids_biman_2024}.  
These activities require coordinated motion of both arms and awareness of nearby objects and surfaces. A third-person camera can capture both arms and the surrounding scene from one viewpoint, which is appealing for vision-based imitation learning~\cite{imitation_survey_2018}. 
Recent imitation learning approaches show that large demonstration datasets yield increasingly general bimanual policies~\cite{black2024pi0visionlanguageactionflowmodel,intelligence2025pi05,liu2024rdt}, but collecting sufficiently large and diverse action-labeled data is costly, which limits scalability. 

Data augmentation has emerged as a promising strategy to address this bottleneck, particularly in single-arm manipulation, where novel views and corresponding action labels can be synthesized offline~\cite{tian2024vista}. 
However, extending these techniques to bimanual manipulation, depth images, and beyond wrist-mounted (eye-in-hand) viewpoints~\cite{liu2025DCODA,zhang2024diffusionmeetsdagger}, introduces new difficulties. These include enforcing scene-wide visual consistency across two robot arms, preserving action label correctness, and handling the higher degrees-of-freedom (DOF). 

In this paper, we propose Synthetic \textbf{\underline{Ro}}bot \textbf{\underline{P}}ose Generation for RGB-D Bimanual Data \textbf{\underline{A}}ugmentation (\method), a data augmentation method for vision-based imitation learning of bimanual manipulation. Our key insight in \method is to adapt Pose-Guided Person Image Synthesis techniques~\cite{ma2018poseguidedpersonimage} (originally developed for human pose generation) by conditioning a diffusion model on robot joint configurations. This enables the diffusion model to yield realistic and consistent third-person views for varied robot arm poses. We maintain a tight correspondence between augmented visual observations and valid actions, so that policies trained on the augmented dataset remain executable on real hardware.  

We evaluate \method across 5 simulated and 3 real-world bimanual manipulation tasks, based on the PerAct2~\cite{peract2} benchmark in simulation and a physical bimanual UR5 setup. See Figure~\ref{fig:teaser} for an example rollout. Compared to an Action Chunking with Transformers (ACT)~\cite{Zhao-RSS-23} baseline trained only on raw demonstrations, data augmentation using \method enables ACT to achieve higher success rates across bimanual manipulation tasks that require coordinated motion and fine-grained precision. 
By enabling third-person vision-based data augmentation for bimanual systems, \method scales training data and broadens coverage without extra manual data collection.

The contributions of this paper include:
\begin{itemize}
    \item \method, a novel pose-guided image synthesis method for offline data augmentation in bimanual manipulation, which supports learning from both RGB and RGB-D data. 
     \item Depth image synthesis that generates depth maps consistent with the augmented joint positions and RGB images.
    \item Action-consistent augmentation that outputs images and joint-space labels, with constraints to ensure feasibility. 
    \item Simulation and real-world experiments showing that bimanual policies with \method achieve significantly improved performance over baseline methods and ablations. 
\end{itemize}

\section{Related Work}

\subsection{Bimanual Manipulation Methods}

Compared to single-arm setups, bimanual systems introduce significant complexity due to larger degrees of freedom and coordination requirements. Classical approaches rely on geometric methods~\cite{maitin2010cloth} or constraint-based representations~\cite{ureche2018constraints}, but assume strong task-specific priors that limit generalization.
This has motivated learning-based methods using reinforcement learning and imitation learning. 
While reinforcement learning shows progress~\cite{BiDexHands2022,robopianist2023,Chitnis2020,Biman_AC_2023}, scaling to robust real-world bimanual policies remains difficult due to high-dimensional control and heavy environment interaction requirements~\cite{chernyadev2024bigym}. 

Imitation learning has therefore emerged as a dominant strategy for ``robotic foundation model'' manipulation systems~\cite{black2024pi0visionlanguageactionflowmodel,intelligence2025pi05}, since demonstrations directly capture desired patterns without exploration~\cite{imitation_survey_2018}. Advances in algorithms~\cite{chi2023diffusionpolicy,Zhao-RSS-23}, hardware~\cite{fu2024mobile,zhao2024alohaunleashed,ding2024bunnyvisionpro}, and datasets~\cite{open_x_embodiment_rt_x_2023} have accelerated this trend.
Despite this progress, data scalability remains a bottleneck. Our contribution is orthogonal to advances in teleoperation and algorithm design; we study imitation learning from third-person views where data augmentation improves robustness without requiring new demonstrations.

\subsection{Data Augmentation for Imitation Learning}
Data augmentation synthetically expands training data at lower cost than collecting new data. In vision classification, simple transformations (e.g., crops, jitters) with preserved labels improve generalization~\cite{krizhevsky2012alexnet}. However, imitation learning suffers from  error accumulation, driving policies into unseen states~\cite{ross2011dagger,Laskey2017DARTNI} which motivates augmentation methods specifically for robotics.

Visual augmentation approaches include altering visual context through structured scene augmentations~\cite{chen2024semanticallycontrollable,yuan2025roboengineplugandplayrobotdata,bharadhwaj2024roboagent,yu2023scaling}, changing backgrounds or objects while keeping robot actions fixed. In contrast, we augment robot poses. Other approaches leverage 3D Gaussian Splatting~\cite{robosplat} but require high-fidelity scene reconstruction, or focus on state-based inputs~\cite{mitrano2022dataaugmentationmanipulation,ke2024ccil} and wrist-camera observations~\cite{zhou2023nerfpalmhand,zhang2024diffusionmeetsdagger,liu2025DCODA}. We study third-person image-based learning and generate full robot poses for richer environmental context.

Unlike approaches that augment new robot states and actions through simulation rollouts~\cite{mandlekar2023mimicgen,jiang2025dexmimicen}, our method operates offline without requiring interactive data collection. Recent diffusion-based methods like VISTA~\cite{tian2024vista} and RoVi-Aug~\cite{chen2024roviaug} enable novel-view synthesis and cross-embodiment transfer. However, these approaches do not generate new action labels and have not been evaluated on bimanual systems. While RoVi-Aug generates multiple views of the same robot pose, \method generates multiple  robot poses from up to four camera viewpoints. Finally, there is relatively limited work studying depth augmentation for robotics. A recent effort~\cite{wang2024visual} uses standard image augmentation techniques to RGB-D images but does not generate new robot poses or action labels.

\subsection{Pose-Guided Image Synthesis}
Pose-guided image synthesis generates realistic images of people in target poses while preserving identity and background context~\cite{ma2018poseguidedpersonimage}. Recent advances leverage diffusion models for higher fidelity and controllability~\cite{bhunia2023personimagesynthesisdenoising,lu2024coarsetofinelatentdiffusion}.
Diffusion-based pose generation has also begun appearing in robotics. GENIMA~\cite{shridhar2024generative} fine-tunes Stable Diffusion with ControlNet~\cite{zhang2023adding} to visualize future joint-action targets as images, then translates them into executable actions. Other approaches employ ControlNet for temporal consistency during action sequence generation~\cite{liu2024diff}. However, these methods focus on visuomotor control rather than generating image-action pairs for imitation learning data augmentation.
Existing works explore differentiable robot rendering~\cite{liu2024differentiablerobotrendering} using 3D Gaussian Splatting, but such approaches~\cite{labbe2021robopose} focus on robot pose reconstruction from images and videos rather than pose generation.
We build on ControlNet's flexibility for guiding diffusion models with structural cues and apply it to bimanual manipulation with skeleton-based conditioning to produce realistic and consistent images.

\begin{figure*}[t]
    \centering
\includegraphics[width=1.0\textwidth]{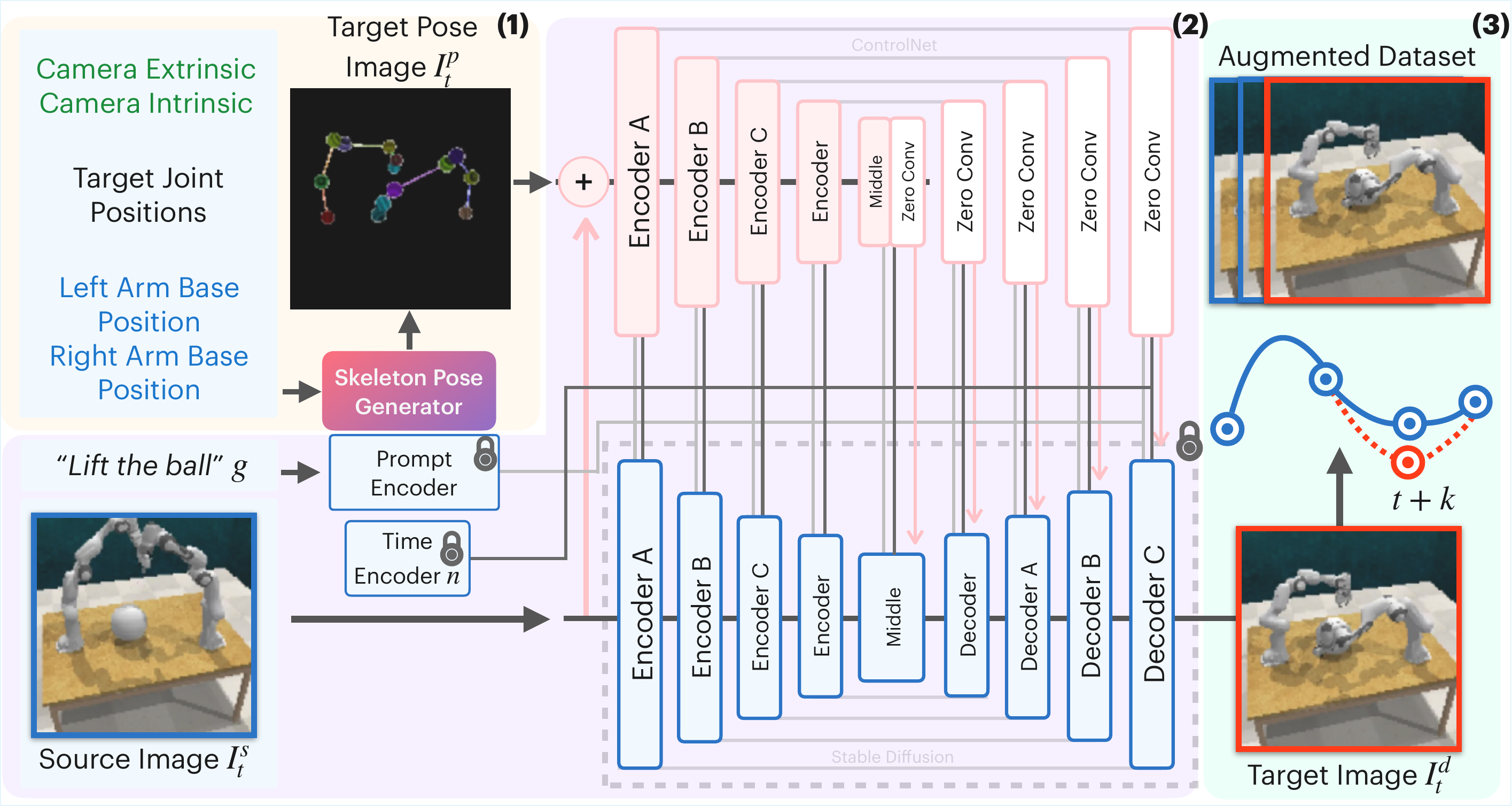}
    \caption{
        \textbf{\method Overview.} \textbf{(1)} The Skeleton Pose Generator takes camera extrinsics and intrinsics, target joint positions, and left and right robot base positions to generate a skeleton pose image $I_{t}^p$ representing the target joint configuration. \textbf{(2)} The source image $I_t^s$ and language goal $g$ are fed into Stable Diffusion (the bottom U-Net model), while the generated skeleton pose serves as control input to ControlNet (the top U-Net model), producing the target image $I_t^{d}$. The locked icons represent frozen parameters. \textbf{(3)} The original dataset is duplicated and \textcolor{figlightred}{generated target states} replace the original states every $k$ timesteps (see Section~\ref{ssec:action-labeling-and-dataset-construction} for more details), with updated corresponding action labels. This augmented dataset is combined with the \textcolor{figdarkblue}{original dataset} to train a bimanual manipulation policy.
    }
    \vspace{-10pt}
    \label{fig:pipeline}
\end{figure*}

\section{Problem Statement}

Our focus is on vision-based, eye-to-hand imitation learning for bimanual manipulation, where a policy $\pi_\theta$ with parameters $\theta$ is trained from expert demonstration data using third-person RGB or RGB-D camera images.

We denote the source camera image at time ${t}$ by $I_{t}^{s}$ (where $s$ denotes source) and the target camera image by $I_t^d$ (where $d$ denotes target), with the camera intrinsic parameters and extrinsic parameters (camera pose relative to world frame). 
We also denote the left and right robot arms using superscripts $l$ and $r$, and indicate their base positions relative to world frame as $b^l$ and $b^r$, respectively.
The source camera image serves as the policy input which produces actions ${\ba_t = \pi_\theta(I_t^s)}$. The action ${\ba_t = (\ba_t^l, \ba_t^r)}$ specifies target joint positions for the left and right arms from the policy.
Training relies on a dataset of $M$ expert demonstrations ${\mathcal{D} = \{\tau_1, \ldots, \tau_M\}}$, where each $\tau_i$ is a sequence of third person camera image observations and corresponding actions: ${\tau_i = (I_1^s, \ba_{1}^l, \ba_{1}^r, \ldots, I_T^s, \ba_{T}^l, \ba_{T}^r)}$ for a demonstration with $T$ time steps. 

\section{Method: \method}

\method leverages a conditional diffusion model to generate novel robot poses while preserving scene context. The diffusion model takes in a source camera image ${I_{t}^s}$, target skeleton pose image $I_{t}^p$, and language goal ${g}$ as input, and outputs the target camera image ${I_{t}^{d}}$. The key objective is to translate a source robot image into a desired target pose. This is accomplished through a three-stage process: fine-tuning Stable Diffusion on expert demonstrations (Section~\ref{ssec:pgris}), generating the skeleton pose for the new robot poses (Section~\ref{ssec:skeleton-pose-generator}), and generating action labels to synthesize new robot poses and reconstructing the dataset (Section~\ref{ssec:action-labeling-and-dataset-construction}). Figure~\ref{fig:pipeline} provides an overview.

\begin{figure*}[t]
    \centering
    \includegraphics[width=1.0\textwidth]{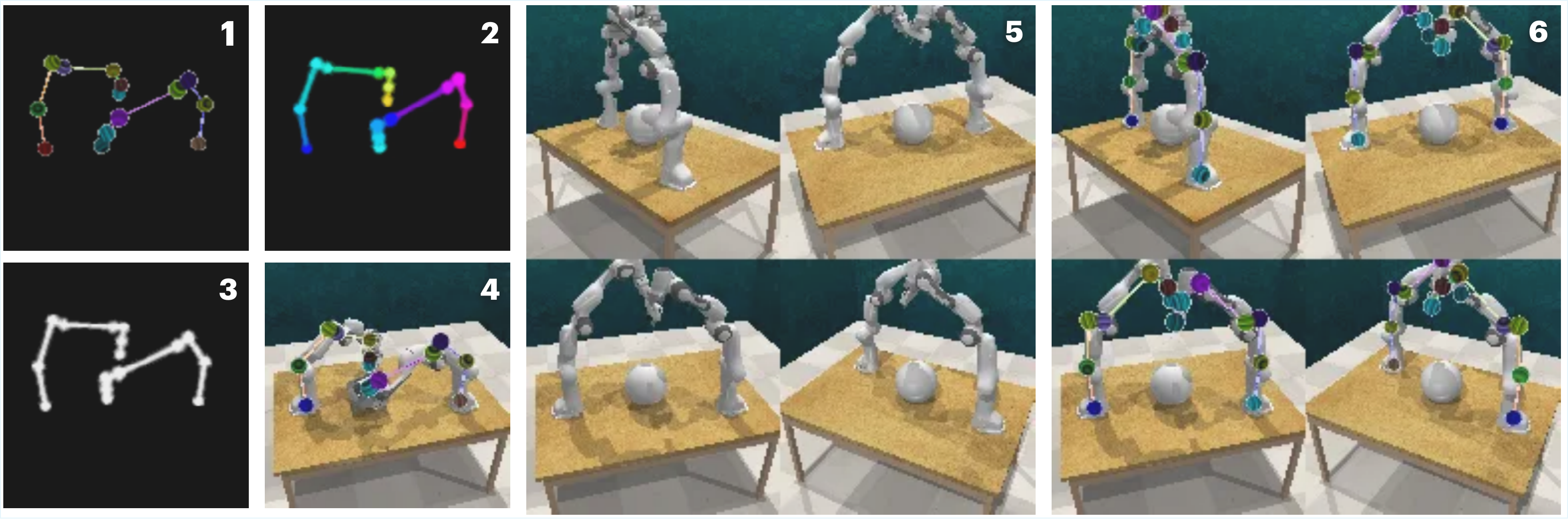}
    \caption{
        \textbf{Skeleton Pose Ablations and Visualization.}
        Comparison of different skeleton pose formats: \textbf{(1)} \method's skeleton pose, \textbf{(2)} OpenPose~\cite{Cao2019OpenPose} inspired skeleton pose, and \textbf{(3)} an all white Skeleton Pose (less visual contrast). \textbf{(4)} demonstrates precise alignment between \method's skeleton pose and the source image. \textbf{(5)} shows a source image input for multi-view generation, while \textbf{(6)} displays the skeleton pose for both robots overlaid on that same source image.
    }
    \label{fig:skeleton_pose_ablations}
    \vspace{-10pt}
\end{figure*}

\subsection{Pose Guided Robot Image Synthesis (PGRIS)}
\label{ssec:pgris}
Inspired by Pose Guided Person Image Synthesis (PGPIS)~\cite{ma2018poseguidedpersonimage}, we formulate the problem of \textbf{Pose Guided Robot Image Synthesis (PGRIS)}. PGRIS aims to synthesize novel robot configurations while maintaining environmental consistency. Formally, given a source camera image $I_t^s$, a target skeleton pose image $I_{t}^p$ (described in Section~\ref{ssec:skeleton-pose-generator}), and a language goal $g$, PGRIS seeks to learn a mapping:
\begin{equation}
I_t^d = f_\psi(I_t^s, I_{t}^p, g),
\end{equation}
where $f_\psi$ is a conditional generative model parameterized by $\psi$. 
In our implementation, $f_\psi$ corresponds to a \textbf{Stable Diffusion 2.1}~\cite{rombach2022high} model fine-tuned with ControlNet~\cite{zhang2023adding}. ControlNet employs a two-stream architecture: a frozen Stable Diffusion U-Net that processes noisy latents and text embeddings, and a copy of the Stable Diffusion U-Net that serves as the trainable encoder stream processing conditioning images. The source image $I_t^s$ and target pose image $I_{t}^p$ are encoded into latent representations $z_s = E(I_t^s)$ and $z_{pose} = E(I_{t}^p)$ using the pretrained Stable Diffusion encoder $E$. The language goal $g$ is embedded via a text encoder~\cite{radford2021clip}: $c_{text} = \xi(g)$.

The trainable ControlNet module $\mathcal{C}_\phi$ processes the conditioning inputs $(z_s, z_{pose})$ and injects pose-aware feature maps into the frozen U-Net denoiser $\epsilon_\phi$ through zero-initialized convolution layers~\cite{zhang2023adding}. These zero-convolutions encode spatial constraints that guide the denoising process to the desired robot configuration while preventing harmful noise from affecting the network during early training phases. At training time, Gaussian noise $\epsilon$ is added to the target latent $z_d = E(I_t^d)$ at diffusion step $n$ following the DDPM noise schedule~\cite{ho2020denoisingdiffusionprobabilisticmodels}, producing $z_{d,n} = \sqrt{\alpha_n} z_d + \sqrt{1-\alpha_n} \epsilon$. The model is trained to predict the noise term by minimizing the following loss:
\begin{equation}
\label{eq:noise}
\mathcal{L} = \mathbb{E}_{z_d, z_s, z_{pose}, g, n, \epsilon} \big\| \epsilon - \epsilon_\phi(z_{d,n}, c_{text}, n; \mathcal{C}_\phi(z_s, z_{pose})) \big\|_2^2
\end{equation}
During inference, the model iteratively denoises a random Gaussian latent with ControlNet guidance to produce the final target image $I_t^d = D(z_d^*)$ using DDIM sampling~\cite{song2021ddim}. We train the diffusion model on expert demonstrations $\mathcal{D}$ for a fixed number of iterations, then use the trained model with sampled camera perturbations to synthesize novel views from $\mathcal{D}$.

\subsection{Skeleton Pose Generator}
\label{ssec:skeleton-pose-generator}

In order to train our diffusion model from Section~\ref{ssec:pgris}, we need a scalable method to render large numbers of images showing diverse robot skeleton poses that maintain geometric consistency. Inspired by conventions in human pose estimation frameworks such as OpenPose~\cite{Cao2019OpenPose}, we create a skeletal representation of each arm's kinematic chain using cylinders for bone segments and spheres for joint connectors. This provides explicit geometric structure that enables the diffusion model to understand spatial relationships and maintain kinematic consistency during image generation. We apply striped textures from GENIMA~\cite{shridhar2024generative} to the spheres (see panel 1 in Figure~\ref{fig:skeleton_pose_ablations}) to provide visual cues that help Stable Diffusion learn joint orientations and rotational states for image generation. 
Our approach does not require explicit 3D information because depth-varying spheres and textural cues help encode spatial relationships for the diffusion model. 

To generate the target skeleton pose $I_{t}^p$, we apply an end-effector transformation $\Delta\rho$ (Section~\ref{ssec:action-labeling-and-dataset-construction} shows how we generate the transformation) to the source end-effector poses and solve inverse kinematics to obtain the target joint positions. We then configure the virtual arms to these joint positions and render the scene using the configured virtual camera, capturing the skeleton pose image with identical dimensions $H \times W$ as the source camera image $I_t^s$.

To achieve this, we leverage PyRender~\cite{pyrender}, a Python-based 3D rendering library, to create skeletal visualizations of the robot arms. We configure PyRender's virtual camera to match the intrinsic and extrinsic parameters from the source camera that captured $I_t^s$, ensuring geometric consistency between the rendered skeleton and the real camera view (see panel 4 in Figure~\ref{fig:skeleton_pose_ablations}). We load URDF models of both robot arms in the virtual environment and position them at their respective base locations $b^l$ and $b^r$ to match the source camera image setup.

\subsection{Action Labeling and Dataset Construction}
\label{ssec:action-labeling-and-dataset-construction}
The previous approach enables image synthesis but lacks constraint-enforced action sampling to ensure valid perturbations. To avoid arm collisions or uncoordinated perturbations in coordinated tasks, we leverage camera pose sampling for bimanual manipulation by decomposing tasks into contactless and contact-rich states. For contact detection, we adopt the force-based approach of Zhang et al.~\cite{Zhang2020RobotCollision}, modeling external torque residuals as $\mu_{\text{ext}} = \mu_{\text{motor}} - \hat{\mu}_{\text{model}}$ using an autoregressive model. Contact is detected when actual torque exceeds dynamic thresholds for consecutive timesteps. This approach offers advantages over vision-based detection: (1) it works with third-person cameras where contact points may be obscured, and (2) it avoids the computational overhead of external modules for detection and segmentation (e.g., using a vision-language model for detecting visual contacts). 

We apply pose perturbations $\Delta\rho$ to source joint positions, where $\Delta\rho = {}_{p_s}T_{p_d}$ represents the end-effector transformation from source pose $p_s$ to target pose $p_d$. The joint-space perturbations are computed by solving inverse kinematics for the transformed end-effector poses. The perturbation strategy employs uniform sampling for contactless states and constrained optimization for contact phases to ensure kinematically feasible poses while preserving contact relationships. The perturbations $\Delta\rho$ are used to compute perturbed actions $\tilde{\ba}_t = (\tilde{\ba}_t^l, \tilde{\ba}_t^r)$ from the original $\ba_t$, where the perturbed actions correspond to the joint positions required to achieve the transformed end-effector poses. 
We duplicate the original dataset, replacing original states with augmented out-of-distribution states every $k$ timesteps ($t+k$, $t+2k$, etc.) to mitigate compounding errors in behavior cloning policies, where we set $k=8$ based on trials with different $k$ values.
 This process generates an augmented dataset $\tilde{\mathcal{D}}$ of novel images paired with corresponding action labels, which we combine with the original dataset.

\begin{figure}[t]
    \centering
    \includegraphics[width=0.48\textwidth]{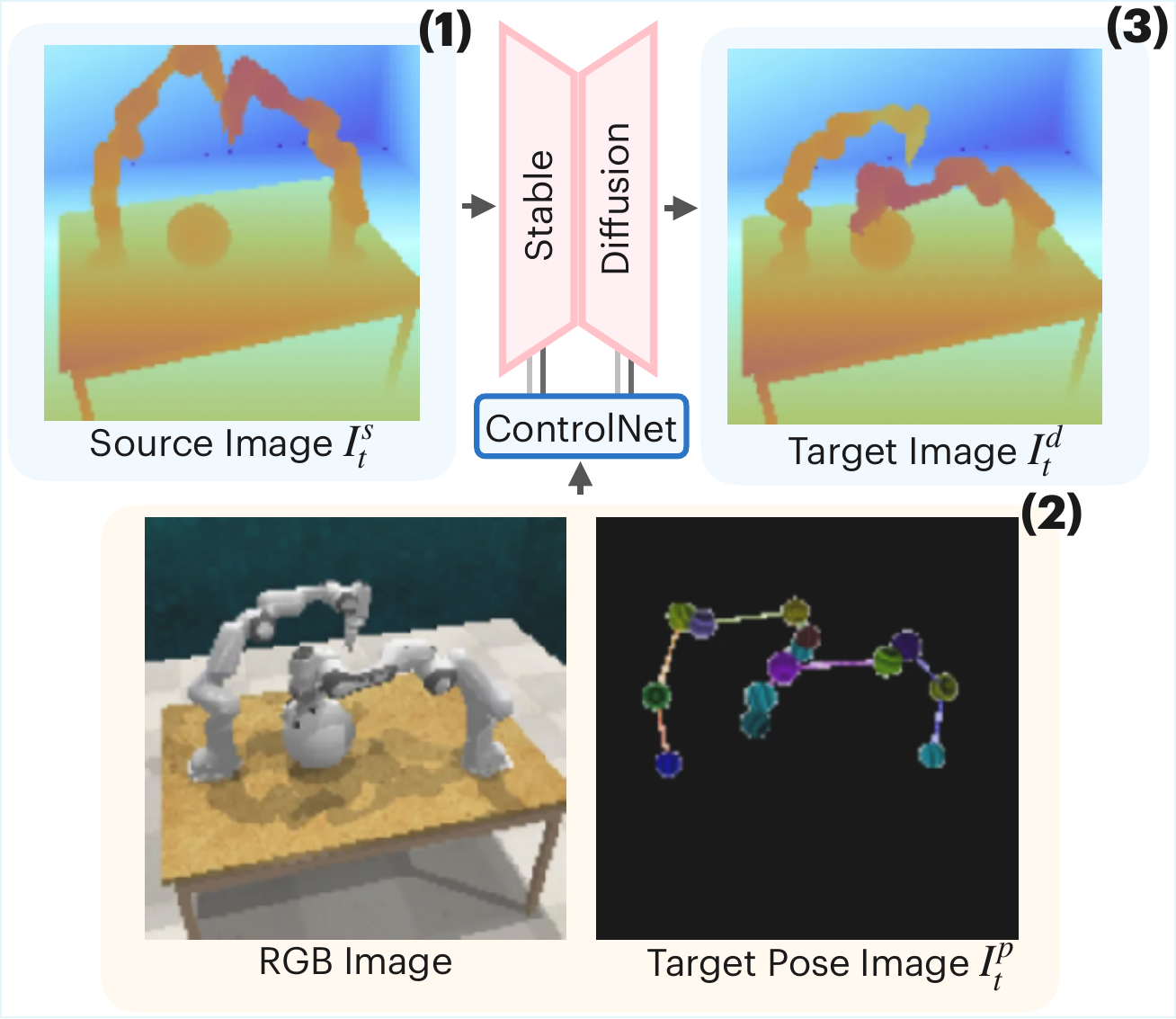}
        \caption{
        \textbf{Depth Image Generation.} Condensed variation of the pipeline in Fig.~\ref{fig:pipeline} for depth image synthesis. \textbf{(1)} Source depth colormap image input to Stable Diffusion. \textbf{(2)} RGB target image and skeleton pose provide conditioning inputs to ControlNet. \textbf{(3)} Generated target depth image.
    }
    \label{fig:depth_image_conversion}
    \vspace{-16pt}
\end{figure}

\subsection{Multi-Modal and Multi-View Extensions}
We extend our approach to incorporate RGB-D data, as depth information is crucial for robot manipulation by providing geometric context and enabling 3D spatial understanding.

For depth image augmentation, we adapt the PGRIS architecture to generate depth-consistent synthetic data. Instead of using an RGB source image $I_t^s$, we input a depth colormap as the source image (see Figure~\ref{fig:depth_image_conversion}, panel 1). To ensure geometric consistency between synthesized RGB and depth modalities, we condition ControlNet on both the corresponding RGB target image and target pose image $I_{t}^p$ (see Figure~\ref{fig:depth_image_conversion}, panel 2). We modify ControlNet to accept this multimodal input by concatenating the two RGB images (6 channels total) as input. This dual conditioning ensures generated depth images maintain geometric correspondence with RGB images (see Figure~\ref{fig:depth_image_conversion}, panel 3). The final augmented dataset contains paired RGB-depth images that are geometrically consistent.

We also extend \method beyond single-camera third-person viewpoints to support multi-camera setups with up to four cameras. Since multiple camera viewpoints are crucial for avoiding occlusions and improving spatial robustness, we address this challenge by tiling four camera observations into a single composite image (see Figure~\ref{fig:skeleton_pose_ablations}, panels 5 and 6). This tiling approach enables multi-view consistent image generation, aligning with GENIMA's~\cite{shridhar2024generative} observations that tiling multiple viewpoints improves consistency. 

\section{Simulation Experiments}

\subsection{Simulation Experiment Setup}
\label{ssec:baselines}
We conduct simulation experiments using the open-source code from PerAct2~\cite{peract2}, which builds upon RLBench~\cite{james2019rlbench} to provide a benchmark for bimanual manipulation. We simplify certain tasks by reducing axes of variation (e.g., constraining the workspace) to enhance the performance of the Action Chunking with Transformers (ACT) baseline~\cite{Zhao-RSS-23}.
We use the following simulation tasks:

\begin{itemize}
\item \texttt{\textbf{Coordinated Lift Ball(CLB)}}: Lift a randomly spawned ball.
\item \texttt{\textbf{Coordinated Lift Tray(CLT)}}: Lift a tray with an item above.
\item \texttt{\textbf{Coordinated Push Box(CPB)}}: Push a large box to a target area.
\item \texttt{\textbf{Bimanual Straighten Rope Easy(BSR)}}: Place both rope endpoints in their target areas.
\item \texttt{\textbf{Coordinated Put Item in Drawer(CPID)}}: Place an item inside one of three specified drawers.
\end{itemize}

We compare \method against these strong baselines:
\begin{itemize}
    \item \textbf{ACT~\cite{Zhao-RSS-23} (w/o Augmentation)}: A state-of-the-art imitation learning method for bimanual manipulation trained on 100 original demonstrations.
    \item \textbf{ACT~\cite{Zhao-RSS-23} (more data)}: Trained on the original dataset along with 100 additional demonstrations without data augmentation, providing a performance reference for ACT with expanded expert data.
    \item \textbf{VISTA~\cite{tian2024vista}}: A diffusion-based novel view synthesis method that uses ZeroNVS~\cite{sargent2024zeronvs} to augment third-person viewpoints from a single third-person view. We extend this to a bimanual manipulation setting.
\end{itemize}

We ensure fair comparisons by using identical training, validation, and test data with consistent environment seeds across all methods. Demonstrations are generated via RLBench's waypoint-based motion planner. 
To maintain consistency with our RGB-only evaluation policy, we adapt ACT to handle both RGB and depth modalities. We modify the input architecture by expanding from 3 to 6 channels. For VISTA, we train ZeroNVS with VISTA's default fine-tuning parameters.

To our knowledge, no prior work has addressed depth-based data augmentation for robotic manipulation. Existing depth estimation models such as DepthAnything~\cite{depthanything} are unsuitable as baselines since they require a set of reference images for fine-tuning, which would be obtained with \method-generated RGB images, but this would create a circular dependency. Consequently, we do not provide baselines for the depth augmentation component of our method.

\begin{figure}[t]
    \centering
    \includegraphics[width=0.48\textwidth]  {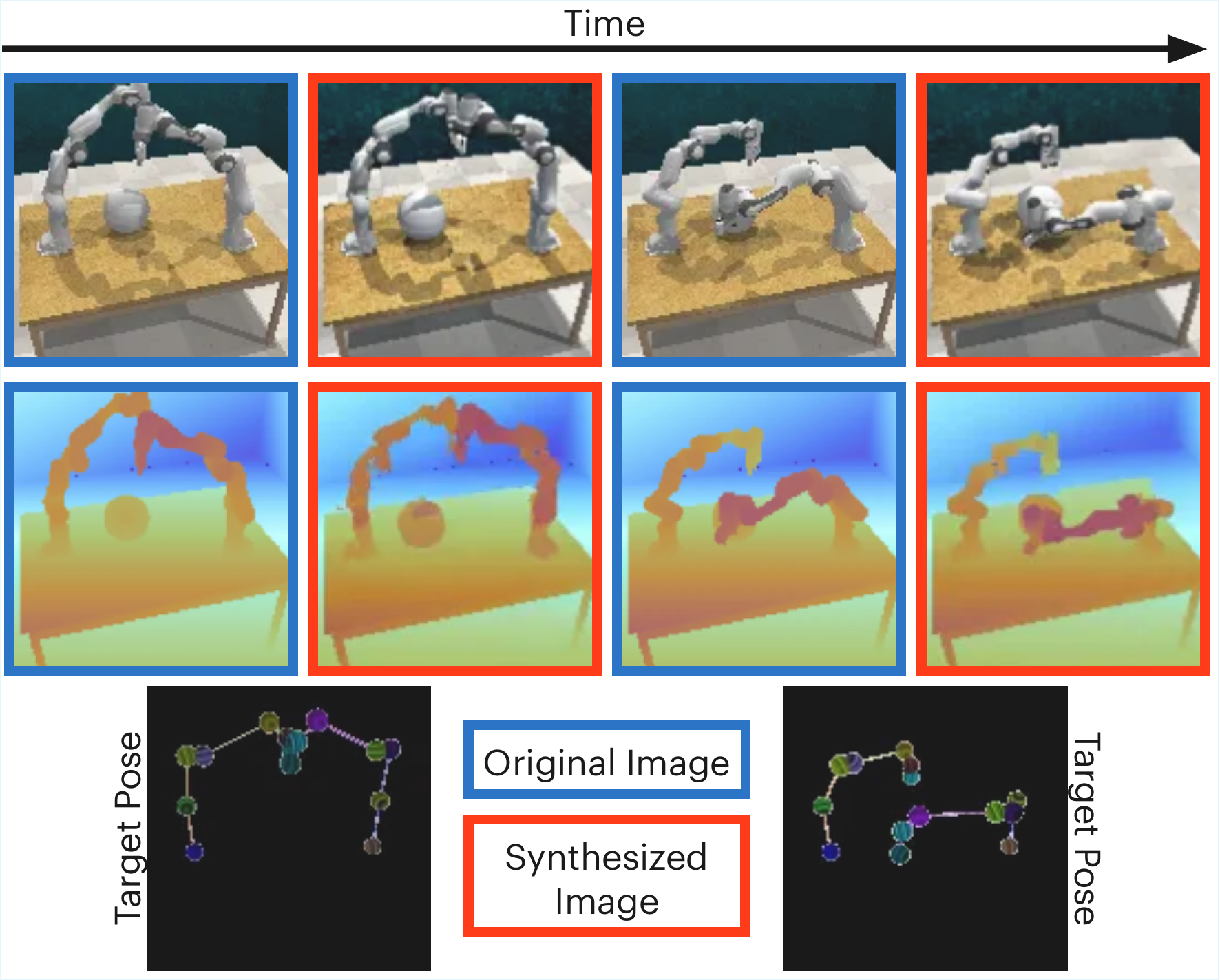}
    \caption{
        \textbf{Synthesized images in simulation.} We present synthesized images from the \texttt{Coordinated Lift Ball (CLB)} task across two timesteps. The \textcolor{figdarkblue}{blue bordered images} show the original RGB and RGB-D images, while the \textcolor{figlightred}{red bordered images} represent the generated target image RGB and RGB-D images conditioned on the corresponding skeleton pose shown below.
    }
    \vspace{-5pt}
    \label{fig:generated_images_simulated}
\end{figure}

\begin{table}
  \setlength\tabcolsep{4.6pt}
  \centering
    \begin{tabular}{lccccccccccc}
    \toprule
        Method & \multicolumn{1}{c}{CLB} & \multicolumn{1}{c}{CLT} & \multicolumn{1}{c}{CPB} & \multicolumn{1}{c}{BSR} & \multicolumn{1}{c}{CPID} \\ 
    \midrule
    RGB ACT (w/o augment.) & 41.3 & 10.7 & 43.3 & 21.3 & 17.3 \\ 
    RGB Fine-tuned VISTA & 52.7 & 1.6 & 54.3 & 10.3 & 15.7 \\
    RGB ACT (more data) & 48.0 & 12.0 & 42.7 & 17.3 & 10.3 \\
    \rowcolor{lightgreen} RGB \method (ours) & \textbf{68.0} & \textbf{30.7} & \textbf{62.7} & \textbf{24.3} & \textbf{30.7}   \\ 
    \midrule
    RGB-D ACT (w/o augment.) & 56.7 & 14.3 & 49.7 & 26.7 & 17.6 \\
    RGB-D ACT (more data) & 33.3 & 9.7 & 47.0 & 20.3 & 16.3 \\
    \rowcolor{lightgreen} RGB-D \method (ours) & \textbf{72.0} & \textbf{35.0} & \textbf{61.3} & \textbf{29.0} & \textbf{35.0}  \\ 
    \bottomrule
    \end{tabular}
    
    \caption{
        \textbf{Simulation Results.} Success rates of \method compared with four baselines, averaged over three random seeds. All methods are evaluated using the ACT policy. The top half of the table shows training with RGB images, while the bottom half shows training with RGB images and depth maps. Task names and baseline details are given in Section~\ref{ssec:baselines}. 
    }
    
  \vspace*{-15pt}
  \label{tab:sim-results}
\end{table} 

\begin{table}[t]
  \setlength\tabcolsep{6pt}
  \centering
    \begin{tabular}{lc}
    \toprule
        Method & CLB \\ 
    \midrule
    One Camera ACT (w/o augment.) & 41.3 \\
    \rowcolor{lightgreen} One Camera \method (ours) & \textbf{68.0} \\
    \midrule
    Four Camera ACT (w/o augment.) & 72.0 \\ 
    \rowcolor{lightgreen} Four Camera \method (ours) & \textbf{80.0} \\ 
    \bottomrule
    \end{tabular}
  \caption{
  \textbf{Multi-Camera Results.} Comparison of single vs. four-camera setups on the Coordinated Lift Ball (CLB) task. Four-camera setup uses tiled observations from four viewpoints. Success rates averaged over three seeds.
  }
  \vspace*{-5pt}
  \label{tab:multi-camera}
\end{table}

\subsection{Simulation Results}

Table~\ref{tab:sim-results} shows the success rates of different methods in simulation, with example synthesized images shown in Figure~\ref{fig:generated_images_simulated}. \method outperforms the baselines on all 5 tasks, achieving substantial improvements across both RGB and RGB-D modalities. Baseline methods, particularly VISTA, struggle with precision-demanding tasks that require low error tolerance, such as grasping trays or manipulating ropes. \method's ability to generate out-of-distribution states enables more robust performance in these failure-prone scenarios, with better error recovery capabilities. Another important finding is that simply increasing training data does not guarantee performance improvements. This highlights the importance of our data augmentation approach, which shows consistent improvements over naive data scaling strategies. We also find that incorporating depth into ACT leads to small performance boost compared to RGB only. By augmenting both the depth and RGB modalities, we see slight but measurable performance improvements across all tasks. Table~\ref{tab:multi-camera} also shows that \method maintains strong performance in multi-camera environments.

\begin{table}[t]
\hspace*{0pt}
\begin{minipage}[t]{0.22\textwidth}
  \setlength\tabcolsep{1pt}
  \centering
    \begin{tabular}{lc}
    \toprule
        Method & CLB \\ 
    \midrule
    Zero-Shot & 50.7 \\
    Few-Shot (10 demos) & 58.3 \\
    Fine-tuned (100 demos) & 68.0 \\ 
    \bottomrule
    \end{tabular}
  \caption{
    \textbf{Generalization Experiments.} Simulation results averaged over 3 seeds.
    }
  \label{tab:generalization-exp}
\end{minipage}%
\hspace{0em}
\begin{minipage}[t]{0.25\textwidth}
  \setlength\tabcolsep{1pt}
    \footnotesize
    \begin{tabular}{lccccccc}
    \toprule
    Method & \multicolumn{1}{c}{CLB} \\ 
    \midrule
    \method with Joint Position Only & 46.7 \\
    \method with White Skeleton & 32.0 \\ 
    \method with OpenPose Skeleton & 62.6 \\ 
    \rowcolor{lightgreen} \method (ours) & \textbf{68.0} \\ 
    \bottomrule
    \end{tabular}
  \caption{
  \textbf{Ablation Experiments.} Simulation results averaged over 3 seeds.
  }
  \label{tab:ablations}
\end{minipage}
\vspace*{-20pt}
\end{table}

\subsection{Generalization Experiments}
We evaluate the diffusion model's ability to generalize to unseen tasks and objects. The model is trained on five PerAct2 tasks: \texttt{Pick Up Notebook}, \texttt{Sweep Dust Pan}, \texttt{Coordinated Push Box}, \texttt{Pick Up Plate}, and \texttt{Coordinated Lift Tray}, then tested on our CLB task using three approaches: \textbf{(1) Zero-Shot} direct image synthesis without fine-tuning, \textbf{(2) Few-Shot} fine-tuned with 10 CLB demonstrations, and \textbf{(3) Fine-tuned} trained on 100 CLB demonstrations directly.
Table~\ref{tab:generalization-exp} shows that direct fine-tuning on the target dataset yields the best performance. However, zero-shot and few-shot approaches achieve reasonable results when target data is limited, though image artifacts increase with fewer training demonstrations.

\subsection{Ablation Studies}
We conduct ablation studies in simulation to address two key questions: \textbf{(1)} How does using Skeleton Pose as input compare to raw joint positions? and \textbf{(2)} How does varying the Skeleton Pose influence image generation quality?

\textbf{Skeleton Pose vs. Raw Joint Positions:} We modify ControlNet to replace Skeleton Pose input with direct 14-dimensional joint position vectors, processed via MLP layers and convolutions to map positions to spatial features. Table~\ref{tab:ablations} shows \method performs best with Skeleton Pose compared to \textit{\method with Joint Position Only}, suggesting skeleton representation provides richer spatial context and kinematic constraints that guide diffusion more effectively than raw positional data.

\textbf{Pose Variation Impact:} We evaluate different skeleton pose representations in Figure \ref{fig:skeleton_pose_ablations}. \textit{\method with White Skeleton} uses an all-white skeleton, causing significant performance degradation by eliminating visual contrast and anatomical structure that guide diffusion, providing minimal conditioning signal. \textit{\method with OpenPose Skeleton} achieves strong performance through clear joint connectivity and structure. However, combining this with GENIMA yields the strongest results because GENIMA's sphere textures provide better visual cues for learning joint orientations and rotational states.

\section{Real-World Experiments}

\begin{figure}[t]
    \centering
    \includegraphics[width=0.48\textwidth]  {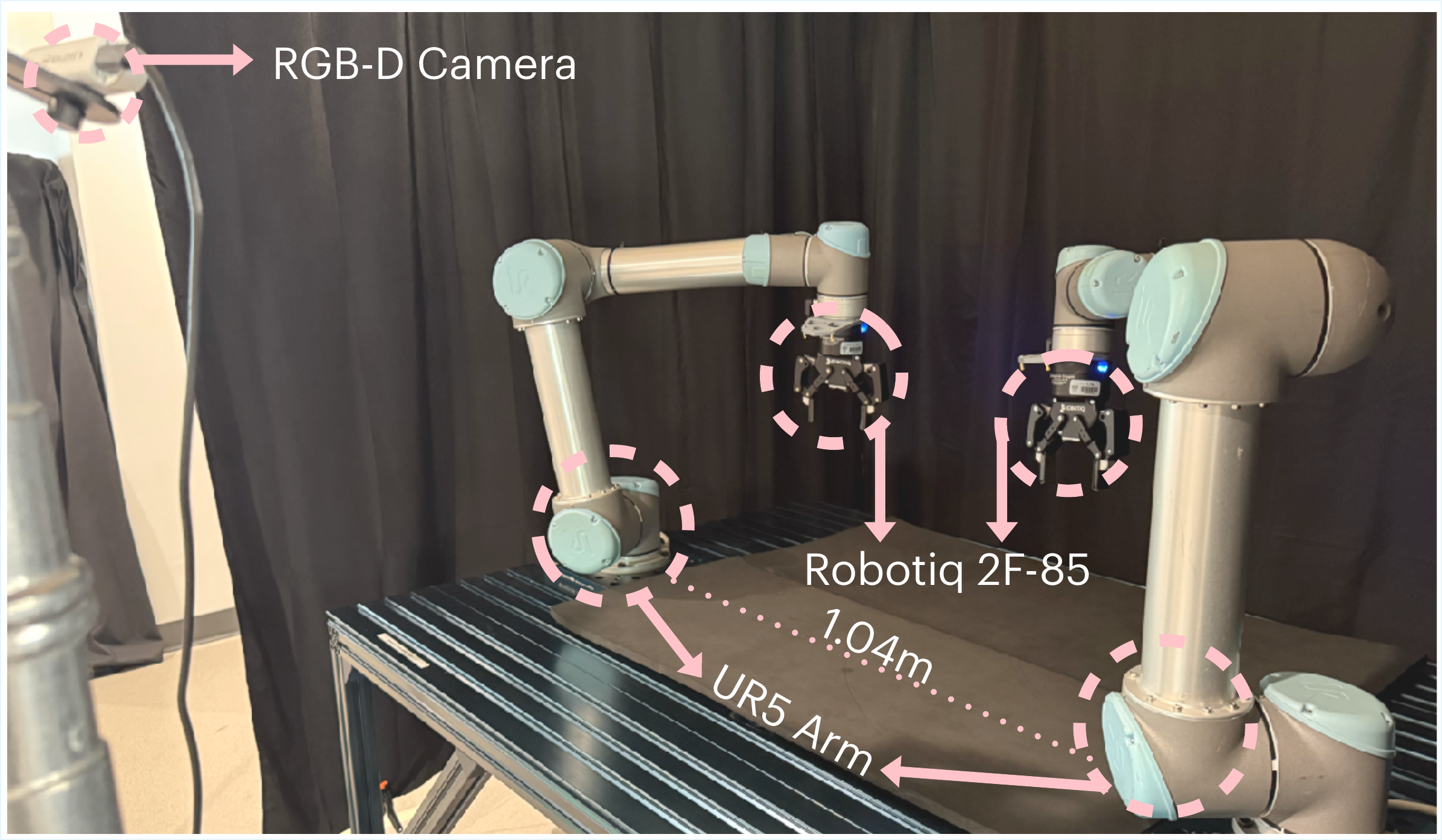}
    \caption{
        \textbf{Real-world setup.} The system features dual UR5 robotic arms in a bimanual configuration, each equipped with a Robotiq 2F-85 gripper. An Intel RealSense D415 RGB-D camera provides visual perception.
    }
    \vspace{-15pt}
    \label{fig:realworld_setup}
\end{figure}

\subsection{Real-World Experiment Setup}

We use two UR5 robots with Robotiq 2F-85 grippers and an Intel RealSense D415 RGB-D camera.
An experienced roboticist collected 30 demonstrations per task using GELLO~\cite{wu2024gello} (see Figure~\ref{fig:realworld_setup}).
We use the following tasks: 

\begin{itemize}
\item \texttt{\textbf{Lift Ball}}: Lift a ball above a region.
\item \texttt{\textbf{Push Block}}: Push a block past the front of a workspace.
\item \texttt{\textbf{Lift Drawer}}: Lift a drawer above a region.
\end{itemize}

\begin{table}[h!]
  \setlength\tabcolsep{3.0pt}
  \centering
    \footnotesize
    \begin{tabular}{lcccccccccc}
    \toprule
        Method & \multicolumn{1}{c}{Lift Ball} & \multicolumn{1}{c}{Lift Drawer} & \multicolumn{1}{c}{Push Block} \\ 
    \midrule
    RGB Fine-tuned VISTA & 11 / 20 & 1 / 20 & 18 / 20 \\
    RGB ACT (w/o augment.) & 13 / 20 & 4 / 20 & 14 / 20 \\ 
    \rowcolor{lightgreen} RGB \method (ours) & \textbf{17 / 20} & \textbf{10 / 20} & \textbf{20 / 20} \\ 
    \midrule
    RGB-D ACT (w/o augment.) & 15 / 20 & 5 / 20 & 17 / 20 \\ 
    \rowcolor{lightgreen} RGB-D \method (ours) & \textbf{19 / 20} & \textbf{13 / 20} & \textbf{20 / 20} \\ 
    \bottomrule
    \end{tabular}
  \caption{
    \textbf{Real-world Experiments.} Results comparing \method with baselines, with 20 trials per method and task combination. 
  }
  \label{tab:real-results}
  \vspace{-10pt}
\end{table} 

\subsection{Real-World Results}
\begin{figure}[t]
    \centering
    \includegraphics[width=0.48\textwidth]  {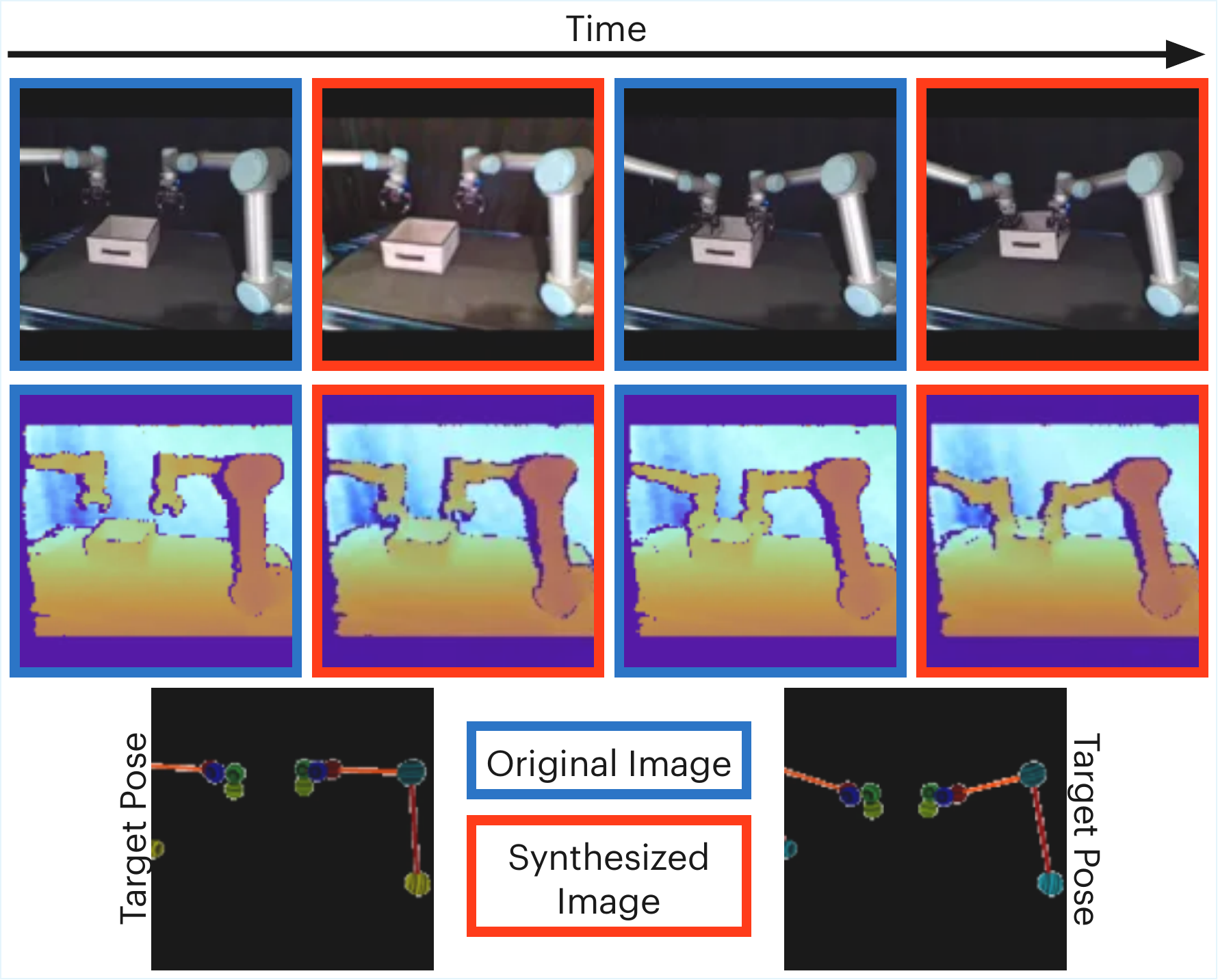}
    \caption{
        \textbf{Synthesized images in the real world.} We present synthesized images from the \texttt{Lift Drawer} task across two timesteps. The \textcolor{figdarkblue}{blue bordered images} show the original RGB and RGB-D images, while the \textcolor{figlightred}{red bordered images} represent the generated target image RGB and RGB-D images conditioned on the corresponding skeleton pose shown below.
    }
    \vspace{-15pt}
    \label{fig:generated_images_realworld}
\end{figure}

Real-world results are presented in Table~\ref{tab:real-results}, with synthesized images in Figure~\ref{fig:generated_images_realworld}. \method demonstrates substantially improved robustness across all tasks compared to baselines.

\textbf{Lift Ball:} Baseline methods frequently freeze upon ball contact, with excessive gripping force sometimes triggering safety limits. Even when successful, baselines often achieve imperfect arm alignment, resulting in loose grasps prone to dropping. \method achieves more reliable and stable grasping.

\textbf{Lift Drawer:} Fine-tuned VISTA struggles with precise manipulation, failing to grasp drawers properly or close grippers entirely. This highlights difficulties that this baseline has with fine motor control requirements.

\textbf{Push Block:} \method demonstrates strong performance when blocks are positioned at greater distances or challenging orientations. We observe that the baseline methods frequently get stuck or fail to initiate contact. This shows how \method improves robustness in manipulation tasks requiring precise control and recovery from suboptimal initial conditions.

\section{Limitations} 
Like all methods, \method has limitations.  The reliance on robot poses for image generation necessitates some observability of the robotic system, potentially reducing effectiveness in scenarios with significant occlusions. Additionally, \method requires camera calibration for skeleton rendering. Like other image generation frameworks, the diffusion model may produce artifacts or poorly synthesized objects, though this could be addressed by incorporating additional conditioning modalities such as detected edges or segmentation masks.

\section{Conclusion}
We present \method, a novel data augmentation approach for bimanual robotic manipulation in imitation learning. Through evaluation in both simulation and real-world settings, we demonstrate consistent improvements over baselines. Our work highlights the potential of pre-trained diffusion models in robotics, mirroring their revolutionary impact on image generation. We hope this research inspires further development of robust data augmentation techniques for robotic manipulation.

\section{Acknowledgments}
Huge thanks to SLURM Lab for discussions and helpful writing feedback.

\bibliographystyle{IEEEtran}
\bibliography{main}

\clearpage
\appendix

\subsection{Task Details}
\begin{figure}[h!]
    \centering
    \includegraphics[width=0.48\textwidth]  {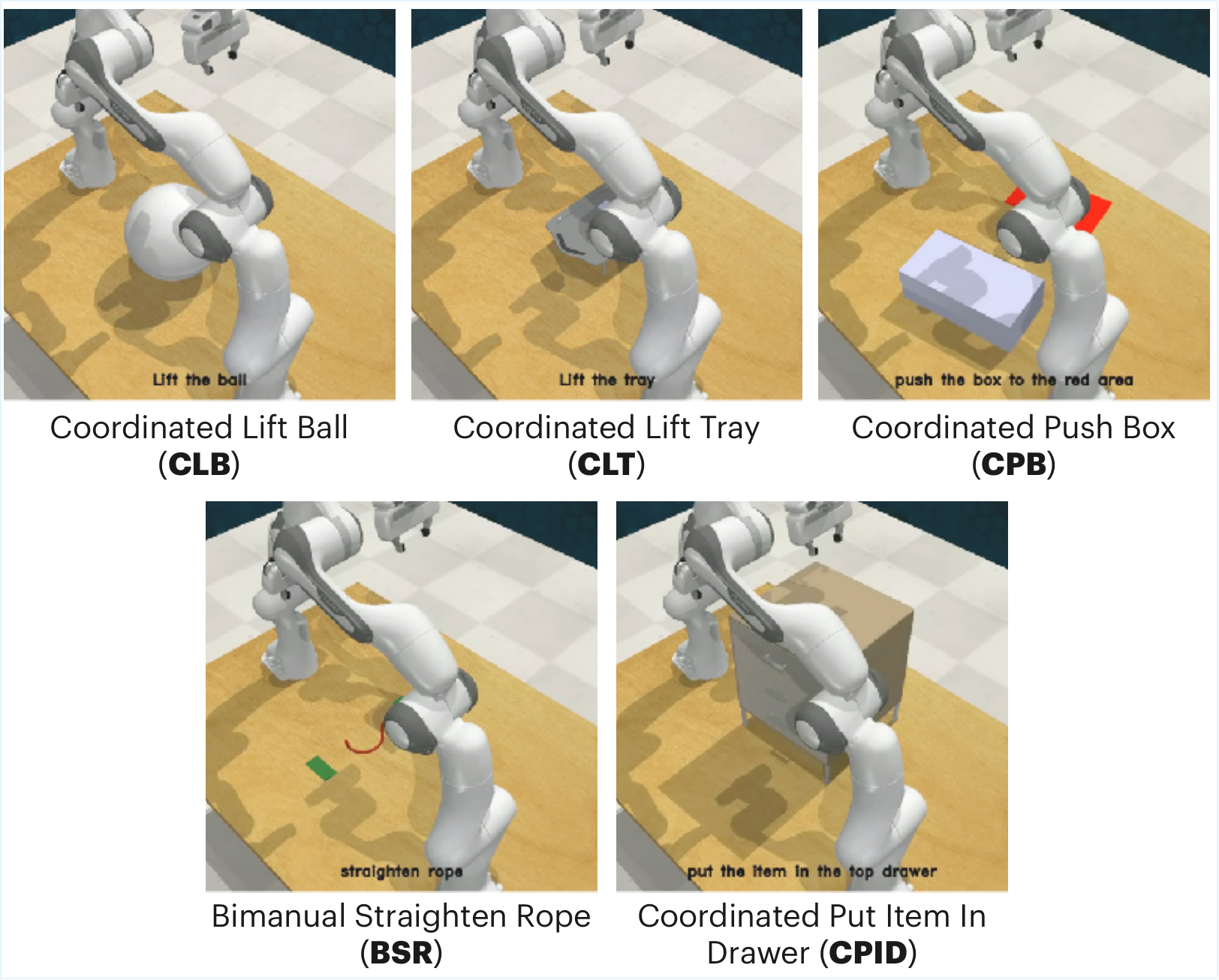}
    \caption{
        \textbf{Simulation environments.} Simulation environment for our bimanual manipulation tasks, adapted from PerAct2. Each simulation image is shown above its corresponding language goal. Text overlays within images indicate the language goal of the task. Abbreviations in parentheses correspond to task names used throughout the paper.
    }
    \vspace{-15pt}
    \label{fig:simulation_tasks}
\end{figure}
\begin{figure}[h!]
    \centering
    \includegraphics[width=0.48\textwidth]  {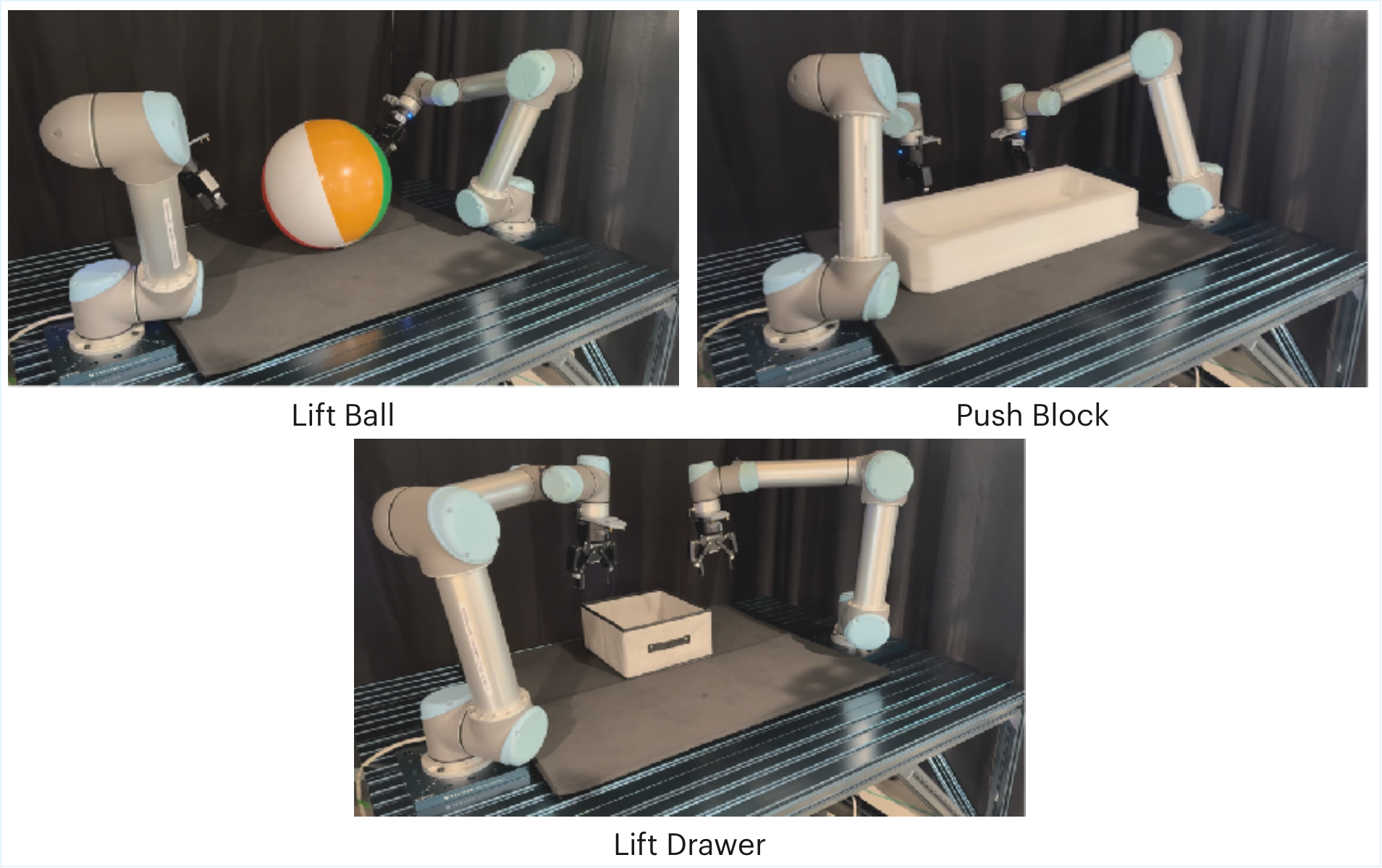}
    \caption{
        \textbf{Real-world environments.} Real-world environment for our bimanual manipulation tasks. Each simulation image is shown above.
    }
    \vspace{-5pt}
    \label{fig:realworld_tasks}
\end{figure}
We show the simulation environment in Figure~\ref{fig:simulation_tasks} and the real-world environment in Figure~\ref{fig:realworld_tasks}. 
For real-world experiments, we use an Intel RealSense D415 camera to capture RGB and RGB-D images at a resolution of $640 \times 480$ pixels. These images are first zero-padded and then rescaled to $128 \times 128$. We use the \href{https://github.com/SintefManufacturing/python-urx}{python-urx} library to control the robot arms and I/O programming to operate the Robotiq 2F-85 grippers.

\subsection{Hyperparameters}
\begin{table}[h!]
  \centering
  \vspace{-2pt}
  \setlength\tabcolsep{5.0pt}
    \footnotesize
    \begin{tabular}{lc}
    \toprule
    Hyperparameter & Value \\ 
    \midrule
    Learning Rate & 1e-5 \\
    Batch Size & 16 \\
    \# Encoder Layers & 4 \\
    \# Decoder Layers & 7 \\
    Feedforward Dimension & 3200 \\
    Hidden Dimension & 512 \\
    \# Heads & 8 \\
    Beta & 100 \\
    Dropout & 0.1 \\
    \bottomrule
    \end{tabular}
  \caption{
    Hyperparameters of ACT.
  }
  \label{tab:hparam-act}
  \vspace{-2pt}
\end{table}
Table~\ref{tab:hparam-act} summarizes the ACT hyperparameters. We use the default PerAct2 chunk size of 10 for all simulation tasks and a chunk size of 2 for all real-world tasks. In both simulation and real-world, the RGB and RGB-D images are $128 \times 128$. An NVIDIA 4090 GPU is used to train the ACT policy.

\begin{table}[t]
  \centering
  \vspace{-2pt}
  \setlength\tabcolsep{5.0pt}
    \footnotesize
    \begin{tabular}{lc}
    \toprule
    Hyperparameter & Value \\ 
    \midrule
    Base Model & Stable Diffusion 2.1 \\
    Learning Rate & 1e-5 \\
    Weight Decay & 1e-2 \\
    Epochs & 150 \\
    Batch Size & 24 \\
    Image Size & $512 \times 512$ \\
    \bottomrule
    \end{tabular}
  \caption{
    Hyperparameters of ControlNet.
  }
  \label{tab:hparam-controlnet}
  \vspace{-2pt}
\end{table}
Table~\ref{tab:hparam-controlnet} summarizes the ControlNet hyperparameters. ControlNet was trained on a single NVIDIA A100 GPU with 80GB of VRAM. Image synthesis was done on a single NVIDIA P100 GPU.

\subsection{Examples of Synthesized Images}
\begin{figure*}[t]
    \centering
    \includegraphics[width=1.0\textwidth]  {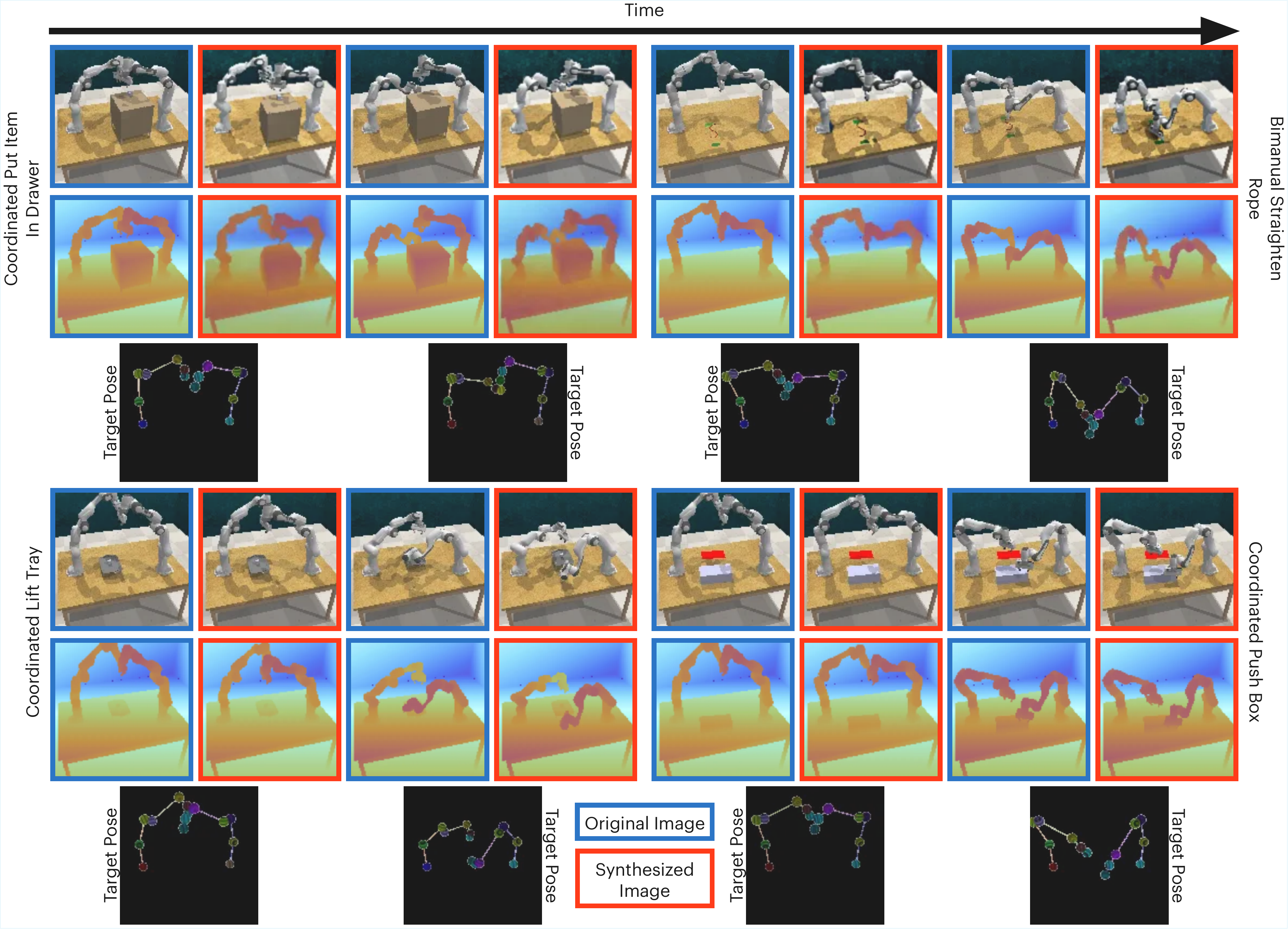}
    \caption{
        \textbf{Synthesized images in simulation.} We present synthesized images from the \texttt{Coordinated Put Item In Drawer (CPID)}, \texttt{Bimanual Straighten Rope (BSR)}, \texttt{Coordinated Lift Tray (CLT)}, and the \texttt{Coordinated Push Box (CPB)} task across two timesteps. The \textcolor{figdarkblue}{blue bordered images} show the original RGB and RGB-D images, while the \textcolor{figlightred}{red bordered images} represent the generated target RGB and RGB-D images conditioned on the corresponding skeleton pose shown below.
    }
    \vspace{-5pt}
    \label{fig:simulation_tasks_all}
\end{figure*}

\begin{figure*}[t]
    \centering
    \includegraphics[width=1.0\textwidth]  {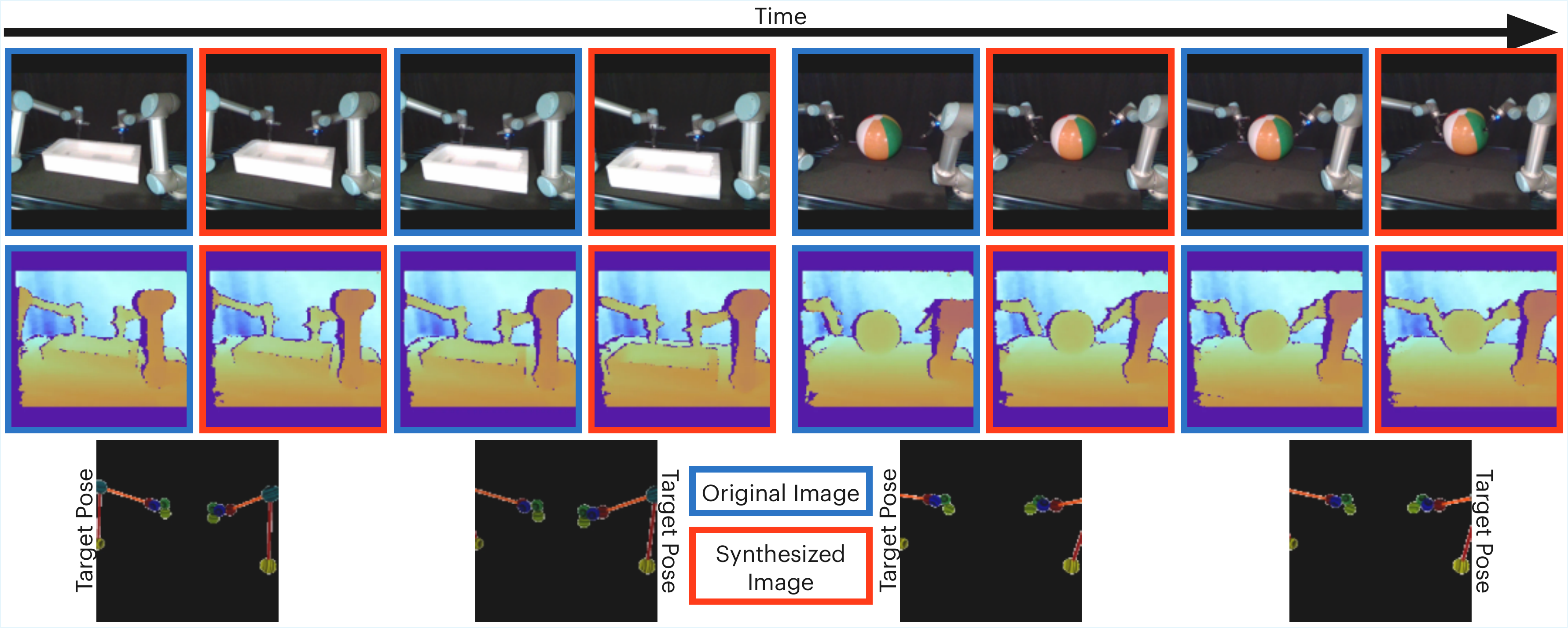}
    \caption{
        \textbf{Synthesized images in the real-world.} We present synthesized images from the \texttt{Push Box} and \texttt{Lift Ball} task across two timesteps. The \textcolor{figdarkblue}{blue bordered images} show the original RGB and RGB-D images, while the \textcolor{figlightred}{red bordered images} represent the generated target RGB and RGB-D images conditioned on the corresponding skeleton pose shown below.
    }
    \vspace{-5pt}
    \label{fig:realworld_tasks_all}
\end{figure*}

Figure~\ref{fig:simulation_tasks_all} shows the simulation synthesized image results for \texttt{Coordinated Put Item In Drawer (CPID)}, \texttt{Bimanual Straighten Rope (BSR)}, \texttt{Coordinated Lift Tray (CLT)}, and the \texttt{Coordinated Push Box (CPB)} tasks. Figure~\ref{fig:realworld_tasks_all} shows the real-world synthesized image results for ~\texttt{Push Box} and ~\texttt{Lift Ball} tasks.

\subsection{Additional Baseline Implementation Details}
For the fine-tuned VISTA approach, we randomly sample 10 overhead camera viewpoints from a quarter-circle arc distribution to train ZeroNVS using VISTA's default fine-tuning hyperparameters. The ZeroNVS model is fine-tuned for 5,000 iterations on four NVIDIA A40 GPUs. The resulting fine-tuned model is used to synthesize overhead camera views for all timesteps across each demonstration episode. These synthetically generated overhead images serve as replacements for the original overhead camera data and are utilized to train the ACT policy.

\subsection{Camera Perturbation Sampling}
For contact-based states, we utilize the constraint optimization framework from D-CODA~\cite{liu2025DCODA} to ensure consistent perturbations across both robotic arms. The approach leverages Dual Annealing~\cite{XIANG1997216}, a global optimization method that handles constrained problems with early termination capabilities. The optimization variable consists of translation coordinates $c_{\text{trans}}$, which define the transformation applied to camera perturbations (normalized within $[-1, 1]$).

The objective function incorporates penalties for several undesirable conditions: perturbations that are too small, end-effector configurations positioned too near the table surface, and end-effector poses too close with the other. To validate the kinematic feasibility of perturbed end-effector positions, we integrate a Levenberg-Marquardt (LM) inverse kinematics solver. Configurations that fail to produce valid joint solutions receive appropriate penalty weights in the optimization process. We define the overall optimization problem as:

\begin{equation}
\begin{aligned}
\underset{c_{\text{trans}}}{\text{minimize}} \quad & \text{Cost}(c_{\text{trans}}) \\
\text{subject to} \quad 
& c_{\text{trans}} \in [-1, 1]^3, \; c_{\text{trans}} \geq m_{\text{lb}}, \\
& \text{ProximityToTable}(c_{\text{trans}}) \geq d_{\text{table}}, \\
& \text{ProximityToOtherEEF}(c_{\text{trans}}) \geq d_{\text{eff}}, \\
& \text{IKSolver}(c_{\text{trans}}) = \text{valid}.
\end{aligned}
\end{equation}

We configure the perturbation parameters as follows: translation magnitudes are bounded by $[m_{\text{lb}}, m_{\text{ub}}] = [0.05, 0.1]$ meters for both contactless and contact-rich scenarios. Rotational perturbations for contactless states are constrained within $[r_{\text{lb}}, r_{\text{ub}}] = [-28.7^\circ, 28.7^\circ]$. The replacement interval parameter $k$, which determines the frequency at which original states are replaced with the synthesized states is set to $k = 8$ across all simulated and real-world experimental tasks.

\subsection{Multi-Conditioning for ControlNet}
To generate depth images consistent with both the RGB image and target pose image (Figure~\ref{fig:depth_image_conversion}), we modified ControlNet to support multi-conditioning modalities. While the native ControlNet code only supports a single conditioning modality, recent works~\cite{controlnet_plus_plus,yang2025dcconctrolnet,zhao2023uni,peng2024controlnext,hu2023cocktail} have explored incorporating multiple conditioning inputs. However, at the time of publication, these approaches either lack publicly available code, have not been evaluated on skeleton pose conditioning (focusing instead on other modalities such as segmentation masks), or require significantly greater computational resources than the standard ControlNet implementation. Usually, these works require 8 GPU's with each at least 48 GB to train. We anticipate that future work can leverage these novel multi-modal conditioning works to further reduce image artifacts and improve generation quality.

\subsection{Saftey Considerations}
Internet pre-trained diffusion models such as Stable Diffusion~\cite{rombach2022high} exhibit harmful biases~\cite{vice2025exploringbias100texttoimage} which could inadvertently influence robot behavior when fine-tuned for manipulation tasks. Robotic systems trained on synthetic data generated by these models should have extensive and thorough safety evaluations before deployment. To mitigate these risks, we recommend implementing safety guidance mechanisms during inference. These include classifiers to detect inappropriate generations and human oversight of synthesized training data.

\end{document}